%% file: main.tex
\documentclass[runningheads]{llncs}
\usepackage[T1]{fontenc}
\usepackage[tight-spacing=true]{siunitx}
\usepackage[utf8]{inputenc}
\usepackage{adjustbox}
\usepackage{amsfonts}
\usepackage{amsmath}
\usepackage{amssymb}
\usepackage{array}
\usepackage{booktabs}
\usepackage{color}
\usepackage{etoolbox}
\usepackage{framed,multirow}
\usepackage{graphicx}
\usepackage{nicefrac}
\usepackage{overpic}
\usepackage{url}
\usepackage{xspace}

\usepackage[breaklinks=true,bookmarks=false]{hyperref}
\usepackage{cleveref}

\input{macro}

\makeatletter
\renewcommand{\paragraph}{%
  \@startsection{paragraph}{4}%
  {\z@}{0em}{-1em}%
  {\normalfont\normalsize\bfseries}%
}
\makeatother

\begin{document}
\title{Unsupervised Intuitive Physics\\ from Visual Observations}
\titlerunning{Unsupervised Intuitive Physics from Visual Observations}
\authorrunning{Ehrhardt et al.}

\author{Sebastien Ehrhardt\inst{1}\textsuperscript{$\star$}\orcidID{0000-0002-6948-079X} \and Aron Monszpart\inst{2,3}\thanks{Authors contributed equally}\orcidID{0000-0003-0579-5300} \and Niloy Mitra\inst{2}\orcidID{0000-0002-2597-0914} \and Andrea Vedaldi\inst{1}\orcidID{0000-0003-1374-2858}}
\institute{University of Oxford, Oxford, UK, \email{\{hyenal, vedaldi\}@robots.ox.ac.uk}\and
 University College London, London, UK,  \email{\{a.monszpart,n.mitra\}@cs.ucl.ac.uk} \and
Niantic, San Fransisco, USA, \email{aron@nianticlabs.com}}
\maketitle

\begin{abstract}
\input{abstract}
\end{abstract}


\input{intro}
\input{relatedWork}
\input{method}

\input{experimentalSetup}

\input{results}
\input{conclusions}

\bibliographystyle{splncs04}

\input{main.bbl}
\end{document}

%% file: macro.tex
\crefname{figure}{Fig.}{Fig.}
\crefname{table}{Table}{Table}

\renewcommand{\paragraph}[1]{\par\smallskip\noindent\textbf{#1}}

\newcommand{\bx}{\mathbf{x}}

\newcommand{\TNet}{\textit{DispNet}\xspace} 
\newcommand{\PTNet}{\textit{PosNet}\xspace} 
\newcommand{\PNet}{\textit{ProbNet}\xspace} 
\newcommand{\IntNet}{\textit{InterpNet}\xspace} 
\newcommand{\netmod}{\textit{Net} models\xspace}

\newcommand{\pool}{\textsc{PoolR}\xspace}
\newcommand{\bowl}{\textsc{BowlR}\xspace}
\newcommand{\heightF}{\textsc{HeightR}\xspace}

\newcommand{\bowls}{\textsc{BowlS}\xspace}
\newcommand{\heightFs}{\textsc{HeightS}\xspace}

\newcommand{\benchmark}{\textsc{Roll4Real}\xspace}
\newcommand{\groundtruth}{ground-truth\xspace}
\newcommand{\heightfield}{height-field\xspace}

\newcommand{\Ttrain}{$T_\text{train}$}

\makeatletter
\DeclareRobustCommand\onedot{\futurelet\@let@token\@onedot}
\def\@onedot{\ifx\@let@token.\else.\null\fi\xspace}

\makeatother

\newcommand{\bowltwo}{\textsc{BowlS}\xspace}

\newcommand{\expl}{\text{Exp.}\xspace}
\newcommand{\impl}{\text{Imp.}\xspace}

\makeatletter
\renewcommand{\paragraph}{%
  \@startsection{paragraph}{4}%
  {\z@}{0.5em}{-1em}%
  {\normalfont\normalsize\bfseries}%
}
\makeatother

%% file: abstract.tex
While learning models of {\em intuitive physics} is an active area of research, current approaches fall short of natural intelligences in one important regard: they require external supervision, such as explicit access to physical states, at training and sometimes even at test time. Some approaches sidestep these requirements by building models on top of handcrafted physical simulators. In both cases, however, methods cannot learn automatically new physical environments and their laws as humans do. In this work, we successfully demonstrate, for the first time,  learning unsupervised predictors of physical states, such as the position of objects in an environment, {\em directly from raw visual observations and without relying on simulators}. We do so in two steps: (i)~we learn to track dynamically-salient objects in videos using causality and equivariance, two non-generative unsupervised learning principles that do not require manual or external supervision. (ii) we demonstrate that the extracted positions are sufficient to successfully train visual motion predictors that can take the underlying environment into account. We validate our predictors on synthetic datasets; then, we introduce a new dataset, \benchmark, consisting of real objects rolling on complex terrains (pool table, elliptical bowl, and random height-field). We show that it is possible to learn reliable object trajectory extrapolators from raw videos alone, without any external supervision and with no more prior knowledge than the choice of a convolutional neural network architecture.

%% file: intro.tex
\section{Introduction}\label{s:intro}

A striking property of natural intelligences is their ability to perform accurate and rapid predictions of physical phenomena using only noisy sensory inputs. Even more remarkable is the fact that such predictors are learned without explicit supervision; rather, natural intelligences induce their internal representation of physics automatically from experience.

Several authors have recently looked into the problem of learning physical predictors using deep neural networks in order to partially mimic this functionality. Early attempts predicted trajectories in hand-crafted spaces of physical parameters, such as positions and velocities, assuming that the \groundtruth values of such parameters are fully observable during training. Others have considered performing predictions from visual observations, but used full supervision for training. Furthermore, while several papers~\cite{chang2016compositional,battaglia2016interaction} make use of simulators as a way to generate the required supervisory signals, limited work has been done in transferring such models to real data.

In this paper, we also investigate learning \emph{physical predictors} using deep neural network. However, we do so in a \textbf{fully unsupervised manner}, learning from observations of unlabelled video sequences. In contrast to approaches such as the recent de-animation method of~\cite{NIPS2017_6620}, we do not require synthetic data, nor do we rely on any  handcrafted physical simulator for prediction. Our models are built directly from real data and learn intuitive physics models that empirically outperform more principled, but more brittle, models based on physical parameters~\cite{sanborn2013reconciling}.

Importantly, our goal is not to merely predict future frames in a video, a problem addressed before by several authors~\cite{lee2018savp}. While we also predict future dynamics from a video stream, our goal is {\em not} to estimate appearance changes, but physical quantities such as object positions and velocities. So, where future frame prediction generates an image, our goal is to extract meaningful and actionable physical parameters from the data.

As a working example, we consider video footage of balls rolling on various surfaces, such as pool tables, bowls and random {\heightfield}s. Balls interact with the underlying environment (e.g., roll around obstacles) and among themselves (e.g., collide with each other). For rigorous assessment, in addition to considering several synthetic datasets, we also contribute \textbf{a new public dataset}, \benchmark, containing a large number of such sequences captured in real-life. Methodologically, we  make two  contributions. First, inspired by~\cite{misra2016shuffle}, we show that an object \textbf{detector} can be learned in an \textbf{unsupervised manner} by tuning a convolutional detector to extract tracks that are maximally characteristic of the natural, causal ordering of the frames in a video. Second, we use these trajectories to learn \textbf{visual predictors} that automatically learn an internal representation of physics and can extrapolate the trajectory of the balls more reliably than even supervised approaches such as  Interaction Networks (IN)~\cite{battaglia2016interaction} that use direct measurements of physical parameters.

Note that our goal, similar to other papers in this area, is not to come up with the best possible method for physical prediction. A handcrafted solution heavily engineered to use supervision, off-the-shelf trackers, and/or physical simulators may do better in raw predictive performance (although the task is in fact not simple, particularly as our terrains are complex and somewhat deformable). Rather, we focus on developing machines that can learn such physical predictors from raw input.

Empirically, we show that vision-based models more gracefully handle observation noise compared to approaches such as~\cite{chang2016compositional,battaglia2016interaction} that are learned using physical \groundtruth parameters extracted from simulated scenarios. We also show that the Visual Interaction Network (VIN) of~\cite{neverova2017predicting}, which also propose a vision-based physical predictor, fails to account for the interaction of the objects and their environment, whereas more distributed tensor based approach succeeds.

The rest of the paper is organized as follows. We discuss related work in~\cref{s:related}. We then present the technical details of our approach in~\cref{s:method}. Next, we introduce the new \benchmark data in~\cref{s:phys} and use the latter as well as several existing synthetic datasets to evaluate the approach in~\cref{s:exp}. We summarise our findings in~\cref{s:conclusions}.

%% file: relatedWork.tex
\section{Related Work}\label{s:related}

Existing work in learning physics can be organised according to several axes.

\paragraph{Nature of the Representation of Physics:} A natural way to represent physics is to manually encode every object parameters and physical properties (mass, velocity, positions, etc.). From the earliest approaches~\cite{battaglia2013simulation} this has been widely used to represent physics and propagate it \cite{battaglia2016interaction,chang2016compositional,2018arXiv06,sanchez2018graph}. Some focusing on representing a small subset of physical parameters such as positions and velocities~\cite{Galileo:NIPS:2015,phys101}. 
However, other approaches try to learn an implicit representation of physics, inspired by the success of implicit representation of dynamics \cite{OndruskaAAAI2016,oh2015action,recurrentenv,bhattacharyya2018long}. Implicit physics are usually represented as activations in a deep neural network~\cite{LearningPhysicalPredictor:emmv:2017,NIPS2017_7040,lerer2016learning}. 

\paragraph{Hand-crafted vs Learned Dynamics:} Some approaches~\cite{Galileo:NIPS:2015}, including \\simulation-based ones~\cite{battaglia2013simulation,wu17learning}, use physics by explicitly integrating parameters such as velocities. While this generally require extensive knowledge of the environment and object properties, other methods~\cite{battaglia2016interaction,chang2016compositional,2018arXiv06,sanchez2018graph,LearningPhysicalPredictor:emmv:2017}, integrate parameters of the scenarios through recurrent learnable predictors to make physical long term predictions. 

\paragraph{Physical vs Visual Observations:} Many approaches~\cite{battaglia2016interaction,chang2016compositional} assume direct access to physical quantities such as positions and velocities for prediction. If this first approach enable to make very accurate predictions it is however unlikely that such accuracy can be reached in the real-world. Others~\cite{battaglia2013simulation,lerer2016learning,li2016visual,wu17learning,NIPS2017_7040,finn2016deep,Stewart2016LabelFreeSO,kansky2017schema} take as input one or several frames of a scene to deduce physical properties (intuitive or explicit) or predict the next state of a system. 

\paragraph{Qualitative vs Quantitative Predictions:} While most of the papers discussed above consider \emph{quantitative} predictions such as extrapolating trajectory, others have considered \emph{qualitative} predictions focusing on \emph{intuitive} physics, such as the stability of stacks of objects~\cite{battaglia2013simulation,lerer2016learning,li2016visual}, the likelihood of a scenario \cite{2018arXiv180307616R} or the forces acting behind a scene~\cite{wu17learning}. Other papers are in between, and learn \emph{plausible if not accurate physical predictions}~\cite{CNNFluid2016,jeong2015data,MonszpartEtAl:SMASH:2016}, often for 3D computer graphics.

\paragraph{Nature of the Supervision:} Most approaches are \emph{passive and supervised}, as they are passive observer of physical scenarios and use ground truth information about key physical parameters (positions, velocities, stability) during training. While this approaches require an expensive annotation of data, some work tried to learn from unsupervised data either through active manipulation~\cite{NIPS2016_6113,denil2016learning} or using the laws of physics~\cite{Stewart2016LabelFreeSO}.

\paragraph{Scenarios:} Two favorite scenarios in such experiments are bouncing balls, including billard-like environment~\cite{fragkiadaki16learning}, and block towers~\cite{lerer2016learning}. As a variant,~\cite{NIPS2017_7040} consider balls subject to gravitational pulls, ignoring harder-to-model collisions. Most papers make use of simulated data, with limited validation on real data. A different approach~\cite{Mottaghi_2016_CVPR} is to predict qualitative object forces and trajectories in fully-unconstrained natural images. The approach of~\cite{NIPS2016_6113} considers instead learning from active poking using a real-life robot. In most cases experiments are done on synthetic data. However, approaches such as \cite{Galileo:NIPS:2015,phys101,li2016visual} also used real data;~\cite{phys101} also contributed a dataset of videos of short physical experiments called~\emph{Phys-101}.

We relate to such previous work in that we also make physical predictions of the trajectory of ball-like objects. However, we differ in two significant ways. First, our approach, while using only passive observations, is \emph{fully unsupervised}, and yet competitive if not more accurate than supervised counterparts. In particular, while~\cite{Stewart2016LabelFreeSO,wu17learning} also do not use image labels, they use \emph{a-priori} knowledge of physics for training (a fully-fledged simulator and renderer in the case of~\cite{wu17learning}). Second, we systematically test on \emph{several real-life scenarios}, both in training and testing, using our new \benchmark dataset. Compared to datasets such as~\emph{Phys-101}, ours allows testing long-term ball-rolling prediction in complex scenarios.

%% file: method.tex
\section{Method}\label{s:method}

\input{fig-tracker}

Our goal is to construct a machine that can, given only raw videos and no supervision, learn physical parameters such as the position 
of the objects in the videos as well as proxies to physical laws that allow to predict the evolution of such parameters over time. For this, predicting appearance changes is not sufficient; instead, we decompose the problem in two steps. The first one is a method to discover and learn to extract object positions using as cue the fact that they should have a non-trivial causal dynamics (\cref{sec:tracker}). This tracker scales well to large datasets and is able to detect different type of objects without any further specification. Then, we use the resulting object trajectories to learn visual predictors that can extrapolate the object positions through time, embodying a proxy to the laws of mechanics (\cref{sec:extrap}).

\subsection{Unsupervised Detection and Tracking of Dynamic Objects}\label{sec:tracker}

\paragraph{Single-object Detection.} Let $\bx_t \in \mathbb{R}^{H\times W\times 3}$ be a RGB video frame and assume we are given video sequences $\mathcal{X}=(\bx_1,\dots,\bx_N)$, initially containing a single object moving in an environment, such as a rolling ball. Our goal is to learn a detector function $\Phi(\bx_t)=u_t \in\mathbb{R}^2$ that extracts the 2D position $u_t$ of the moving object at any given time (\cref{fig:arch_tracker}). The challenge is to do so \textit{without} access to any label for supervision or any a-priori information about object shape. 

We start by implementing $\Phi(\bx_t)$ as a shallow Convolutional Neural Network (CNN) that extracts a scalar score $f_v \in \mathbb{R}$ for each image pixel $v\in\Omega = \{1,\dots,H\}\times\{1,\dots,W\}$, resulting in a heat map. This is then normalised to a probability distribution using the softmax operator $s_v = e^{f_v}/\sum_{z\in\Omega} e^{f_z}$ and the location $u$ of the object is obtained as the expected value $u = \sum_v v s_v$~\cite{finn2016deep}.

We learn $\Phi$ by combining two learning principles. The first one is \textbf{causality}. 
Applied to a video sequence, the detector produces a trajectory $\Phi(\mathcal{X}) = (\Phi(\bx_1),\dots,\Phi(\bx_N))$. We expect that, when the detector locks properly on the rolling object, the trajectory is physically plausible (e.g.,\ causal/smooth); at the same time, if the frames are shuffled by a random permutation $\pi$, the resulting trajectory should \emph{not} be plausible anymore. We incorporate this constraint by learning a discriminator network $D(\Phi(\bx_{\pi_1}),\dots,\Phi(\bx_{\pi_5}))$ that, for a subsequence, can distinguish between the natural ordering of the frames and a random shuffle (top row of \cref{fig:arch_tracker}). The permutation $\pi$ is sampled with 50\% probability as a consecutive sequence of 5 frames ($\pi_{i+1}=\pi_{i}+1$, $i=1,\dots,4$) and with 50\% uniformly at random. The discriminator is a 3 layers multi-layer perceptron followed by a sigmoid and the loss
$$
\mathcal{L}_{disc}(D,\pi)
 = \begin{cases}
- \log D, & \pi_{i+1}=\pi_{i}+1,\ i =1,\dots,4,\\
- \log(1-D), & \text{otherwise}.   
 \end{cases}
$$

The second learning principle is~\textbf{equivariance} (cf., ~\cite{thewlis2017unsupervised,novotny2018self}). This principle suggests that, if a transformation $g$ is applied to a frame $\bx_t$ (e.g., a $\pm \pi/2$  rotation), then the output of the detector should change accordingly: $\Phi(g\bx_t) = g\Phi(\bx_t)$. This is implemented as a Siamese branch (bottom row in~\cref{fig:arch_tracker}) extracting 2D positions $\Phi(g\mathcal{X}) = (\Phi(g\bx_1),\dots,\Phi(g\bx_N))$ from the rotated frames and comparing them to the rotated 2D positions extracted from the original frames using the $L^2$ loss: $\mathcal{L}_{siam} = \frac{1}{N}\sum_t \|g^{-1}\Phi(g\bx_t) -\Phi(\bx_t) \|^2$.

Finally, in order to encourage the softmax operator to produce peaky distributions, we minimise the entropy of the resulting distribution $\mathcal{L}_{ent} = - \sum_{v\in\Omega} s_v \log(s_v)$.
The final loss is therefore $\mathcal{L} = \lambda_d \mathcal{L}_{disc}+ \lambda_e \mathcal{L}_{ent} + \lambda_s \mathcal{L}_{siam}$.
In our experiment, $\lambda_d =1$, $\lambda_e = 0.01$, and $\lambda_s=0.001$.

\paragraph{Multi-object Tracking.} We now extend the method from detection of single objects to tracking of multiple objects. In order to do so, the network is fine-tuned to videos containing two or more moving objects of different appearance.

Since the network produces only a single pair of coordinates, it can formally estimate the location of a single object in the image. However, when multiple objects are present, the unsupervised learning process could still converge to an undesirable result, such as predicting the center of mass of several objects combined, or randomly jumping between objects over time. The first is discouraged by the entropy loss which prefers sharp heat map. The second is discouraged by the causality loss, as discontinuous trajectories would not look plausibly ordered and consistent.

In practice, our model learns to track consistently a single object selected at random among the visible ones. Once this is done, in the next iteration, a second object is detected by suppressing (setting to zero) a circular region of radius $r$ around the first object location in the activations $f_v$ immediately preceding the softmax operator, and the process is repeated for further object occurrences. Before the suppression we also add a positive bias to the activations $f_v$ in order to consider the previously detected objects as zero probabilities in the new heatmap. Note that we consider the number of objects as given since it is in itself already a very challenging task that is under active research~\cite{eslami2016attend}. 


\input{fig-twoballs}

\subsection{Trajectory Extrapolation Networks}\label{sec:extrap}

We consider existing network modules for physical prediction. While these modules use external supervision in the original papers, here we apply them to the output of the unsupervised tracker of~\ref{sec:tracker}, hence training such physical extraploators in a \emph{fully unsupervised} manner for the first time.

We experiment in particular with \PTNet, \TNet, and \PNet from~\cite{LearningMechanics:emvm:2017}, configuring them to take as input the first four frames of a sequence and to produce as output the prediction of future object positions. These models learn an implicit representation of physics, which is extrapolated automatically by a recurrent propagation layer and used to extract estimates of the object positions. The difference between the models is that \PTNet regresses positions from state, while \TNet and \PNet regress displacements from state. Furthermore, \PNet produces a probability estimate over trajectories. 

We also consider the \emph{Visual Interaction Network} (VIN) module and its variant \emph{Interaction Network from State} (IFS)~\cite{NIPS2017_7040}. While VIN uses only visual inputs for prediction just like the other networks, IFS works with an explicit state vector of physical parameters, which we set as the stacking of the 2D positions for four past frames which starts with positions extracted from our tracker. Additionally, in the synthetic experiments (first part of \cref{t:one}), IFS uses velocity and in \bowltwo experiments the \groundtruth ellipsoid axes parameters are appended to the state to inform the model of the shape of the ground. IFS and VIN are trained following~\cite{LearningMechanics:emvm:2017}; in particular, this means that VIN uses the same setting as the original paper ($32\times32$ pixels images). 

We also note that while VIN and models from~\cite{LearningMechanics:emvm:2017} have essentially the same core concepts (they consist of a first feature extractor module to extract implicit physical state, a recurrent propagation module to propagate the state, and an extractor module to get desired physical parameters from the state) their main difference resides in the structure of the propagated state. While VIN used a vector state representation, each of \PTNet, \TNet, and \PNet use a tensor representation.

All such models are trained by showing the network four initial frames of a sequence and the output of the unsupervised tracker up to time $T_\text{train} \in \{15, 20\}$ frames. At test time, the networks, which are recurrent, are used to extrapolate the trajectory up to an arbitrary time $T$, also starting from four video frames. We test in particular $T=T_\text{train}$ and $T \gg T_\text{train}$ to assess the generalization capabilities of the models learned by the network.

In addition, for some experiments on single object we also consider \emph{linear} and \emph{quadratic} extrapolators as baselines. In both cases we fit a first (respectively second) order polynomial to the $10$ first positions given as input (hence with a significant advantage compared to the networks which only observe four frames).

%% file: fig-tracker.tex
\begin{figure}[t!]
    \centering
\includegraphics[width=0.9\columnwidth]{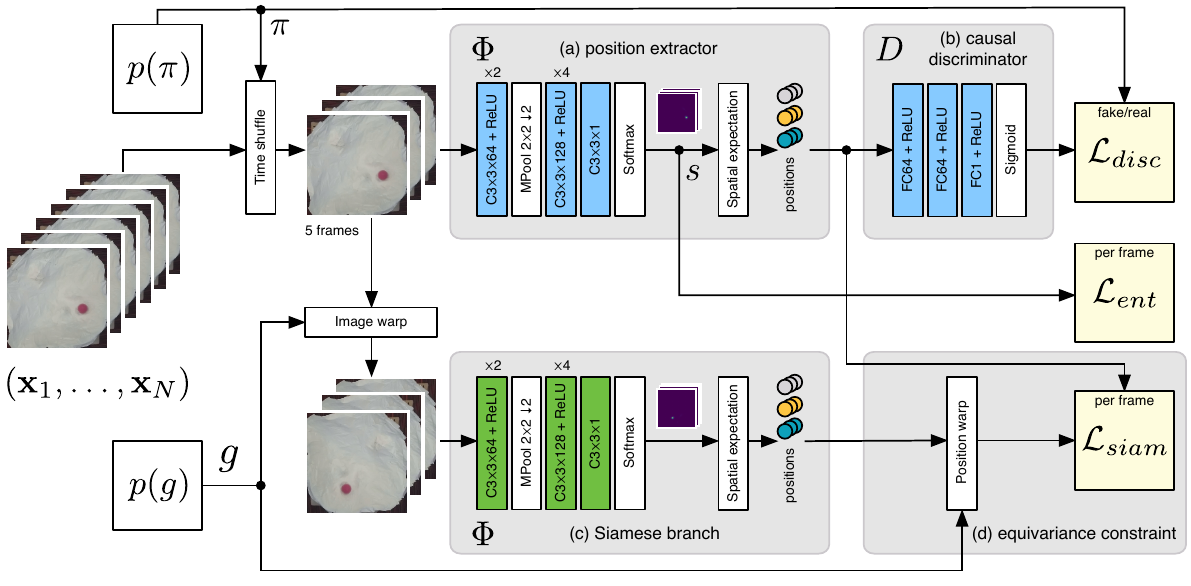}
    \caption{\textbf{Overview of our unsupervised object tracker} Each training point consists of a sequence of five video frames. Top: the sequence is randomly permuted with probability 50\%. The position extractor (a) computes a probability map $s$ for the object location, whose entropy is penalised by $\mathcal{L}_{ent}$. The reconstructed trajectory is then fed to a causal/non-causal discriminator network (b) that determines whether the sequence is causal or not, encouraged by $\mathcal{L}_{disc}$. The bottom Siamese branch (c) of the architecture takes a randomly warped version of the video and is expected by $\mathcal{L}_{siam}$ to extract correspondingly-warped positions in (d). Blue and green blocks contain learnable weights and green blocks are siamese shared ones. At test time only $\Phi$ is retained.}    
    \label{fig:arch_tracker}
\end{figure}

%% file: fig-twoballs.tex
\begin{figure}[t]
    \centering
    \begin{overpic}[width=0.8\linewidth]{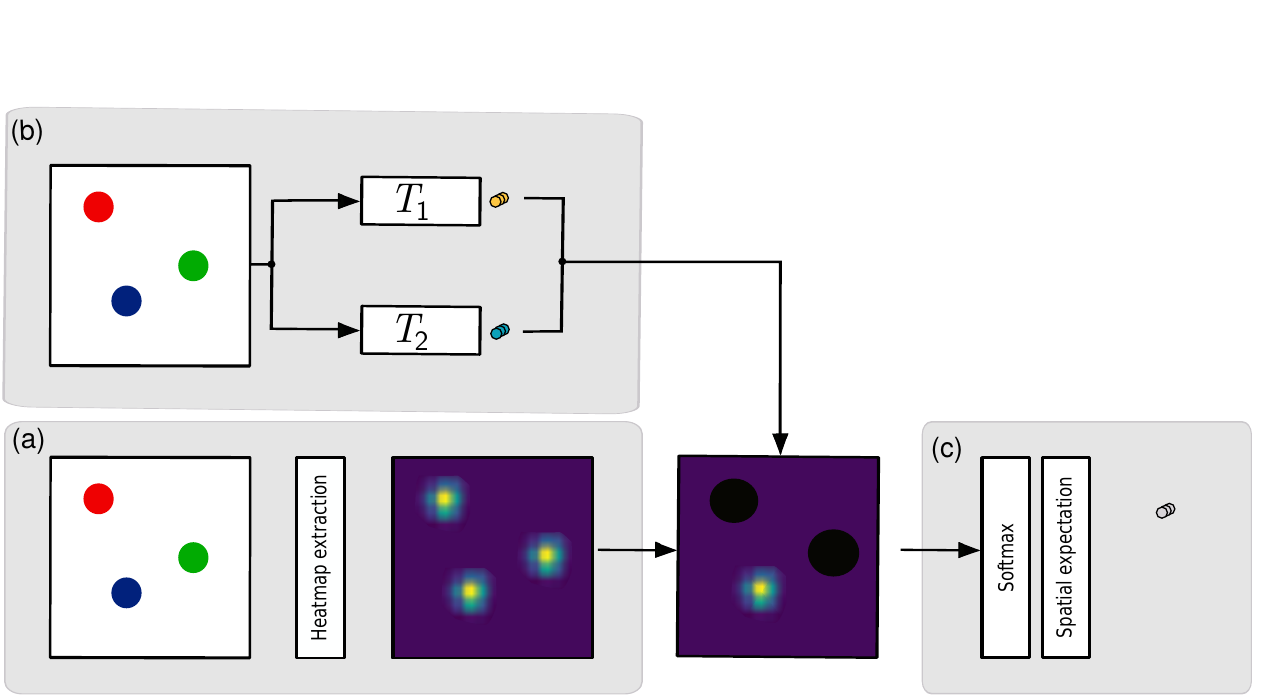}
    \put(87,10){\small $(x_3,y_3)$}
    \end{overpic}
    \caption{\textbf{Multiple object unsupervised tracker} (a) We first extract an object heatmap with the method described in \ref{sec:tracker}. (b) Then we mask the objects detected by previously trained tracker ($T_1$ and $T_2$) on the heatmap by zeroing out the values around a circular area around their center. (c) Finally we extract position from this last heatmap with masked values.}\label{fig:mul_tracker}
\end{figure}

%% file: experimentalSetup.tex
\section{\benchmark: A New Benchmark Dataset}\label{s:phys}

\begin{figure}[t]
    \centering
    \includegraphics[width=\textwidth]{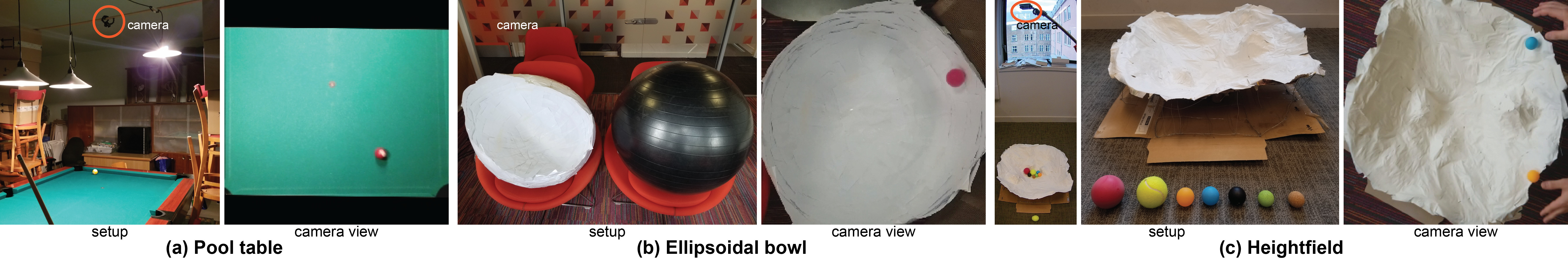}
    \vspace{-2em}
    \caption{\textbf{Physical setup} In each of the three real-world scenarios (\pool, \bowl, \heightF), we show the experimental setup~(left) and a sample data frame~(right).}\label{f:expsetup}
\end{figure}

In the absence of a suitable real-world dataset to evaluate intuitive physics on objects rolling on complex terrains, we created a new benchmark, \benchmark (R4R). 

\paragraph{Dataset Content.} R4R consists of $1118$ short $256 \times 256$ videos containing one or two balls rolling on three types of terrains (\cref{f:expsetup}): a flat pool table (\pool), a large ellipsoidal `bowl' (\bowl), and an irregular \heightfield (\heightF). More specifically, there are
$151$ videos (avg. $99$ frames/video) for the \pool dataset with one ball; 
$216$ videos ($522$ frames/video) for the \bowl dataset with one ball; 
$543$ videos ($356$ frames/video) for the \heightF dataset with one ball; and 
$208$ videos ($206$ frames/video) for the \heightF dataset with two balls.
We rolled a total of 7 differently colored balls for the \heightF and \bowl datasets, varying from \mbox{$3.5$~cm} to \mbox{$7$~cm} in diameter. The \heightfield surface fits into a \mbox{$70\times 70\times 28$~cm$^3$} bounding box, with \mbox{$76$~cm} diameter. The bowl was created using a \mbox{$70$~cm} diameter ball, and is \mbox{$60$~cm} high.
Videos were randomly split into \emph{train}, \emph{validation}, and \emph{test} sets. Ground-truth annotations are provided for the test split.

\paragraph{Dataset Collection.} Both the bowl and \heightfield terrains were modeled using paper m\^{a}ch\'{e} on scaffolds, using a large inflatable ball and a custom-made wire-mesh frame, respectively. For the the \pool dataset, balls were rolled on the table, while for the other settings, balls were manually dropped from a small height and allowed to roll on. The setup was imaged using a fixed camera (Samsung Galaxy S8) from the top. The \pool dataset was captured at 30fps (due to low light), while all the others at 240fps in order to reduce motion blur and later downsampled to 80fps. Videos were cropped to only focus on the scenario of interest, i.e., ball(s) and terrain, and trimmed to retain the portion of the video containing motion.  We rolled a total of 7 different balls: a pink foam ball (\mbox{$7$ cm} diameter), a fluorescent yellow tennis ball (\mbox{$6.8$ cm}), a blue and an orange ping-pong ball (\mbox{$4$ cm}), a black squash ball wit two yellow dots (\mbox{$4$ cm}), and a green and a brown cork ball (\mbox{$3.5$ cm}).

In order to create \groundtruth tracks for the ball centers, we used a template-based tracker using zero-normalized cross-correlation in  the LAB color space, and tracked each frame along with a smoothness term over time. The setup was manually initiated by providing suitable template. The raw results were then manually inspected, corrected, and saved as \groundtruth. We found that due to environment jitter (the ball rolling on the different terrains often created vibration or deformation in the \bowl and \heightF datasets), differences in lighting across some experiments and different ball colors, the template-based tracker was not perfect and manual inspection was required.

It is worth noting that, while this process was enough to produce ground truth annotations for the test set, the method does not scale due to the need for manual verification and correction. While our aim here is to show the feasibility of learning physics in an unsupervised manner, such problems show that our deep tracker also has an applicative advantage compared to these traditional handcrafted approaches.

%% file: results.tex
\section{Results and Discussion}\label{s:exp}

\paragraph{Implementation Details} For all networks trained on every dataset, weights were initialised using Xavier initialization~\cite{glorot2010understanding}. The learning rate was initially set to $10^{-4}$ and was  progressively decreased by a factor of 10 when no improvements were found over $K$ epochs ($K=100$ for the synthetic datasets). Training was stopped when the loss did not decrease for $2K$ consecutive epochs. Before processing images, we resized all dataset images to $128 \times 128$ pixels to fit in the GPU memory. We used TensorFlow~\cite{tensorflow2015:whitepaper} on a single NVIDIA Titan X GPU for all the experiments.

\subsection{Unsupervised Tracker} 
%
%

We first evaluate our unsupervised object detector and tracker and compare against currently state-of-the-art trackers.  We report results in \Cref{t:results} against the following trackers: 1. Optical Flow Lucas-Kanade (OFLK) from OpenCV\cite{opencv_library} library; 
2a. Flownet2-simple, which computes pairwise flowfields using FlowNet2~\cite{ilg2017flownet} and follows the velocity vectors;
2b. Flownet2-blob, where we after computing the flowfields from FlowNet2~\cite{ilg2017flownet}, update the positions as the center of the blobs found in the flowfield. If no blob was detected, we updated the position according to 2a; 3. LAB: a template tracker similar to \cref{s:phys} without any manual corrections.
Note that these methods need manual initialization at the objects positions (expect for LAB) or templating which needs more work with growing object count and/or variety. 
In addition to~\pool, \bowl, and \heightF from \benchmark, we also consider two synthetic datasets from~\cite{LearningMechanics:emvm:2017} in \Cref{fig:bars-tracker}: \bowltwo for the ellipsoidal bowl with one or two balls and \heightFs for the random height-fields. \Cref{fig:bars-tracker}-left reports the mean and $99^{\text{th}}$ percentile pixel error of the extracted object positions against \groundtruth averaged over multiple runs of our experiments.
\input{fig-bartracker}
\input{tab-comp-tracker.tex}
Even though the trackers perform well in practice, they suffer from \emph{large variance}. For example, OFLK went off-track 15\% of the time on the \bowl dataset, 10\% for the \heightF, and 30\% for \pool. In contrast, ours never loses track of the object. The 99th percentile reported in \cref{fig:bars-tracker} shows that the offset is almost constant generally due to the detection occurring on the edge of the objects. Overall, our method learns to track objects robustly in a diverse range of complex scenarios.

Importantly, since our tracker does not use any manual annotations it scales easily to larger synthetic datasets, multiple objects, and different object appearances within the same dataset by just providing more example data.

We also conducted an ablation study on the \bowl dataset to measure the impact of each loss term. \cref{fig:bars-tracker}-right shows that, while each loss contributes to the final results, the best performance is obtained when all the terms are combined. 

\begin{figure}[t!]
    \centering
    \includegraphics[width=.9\columnwidth]{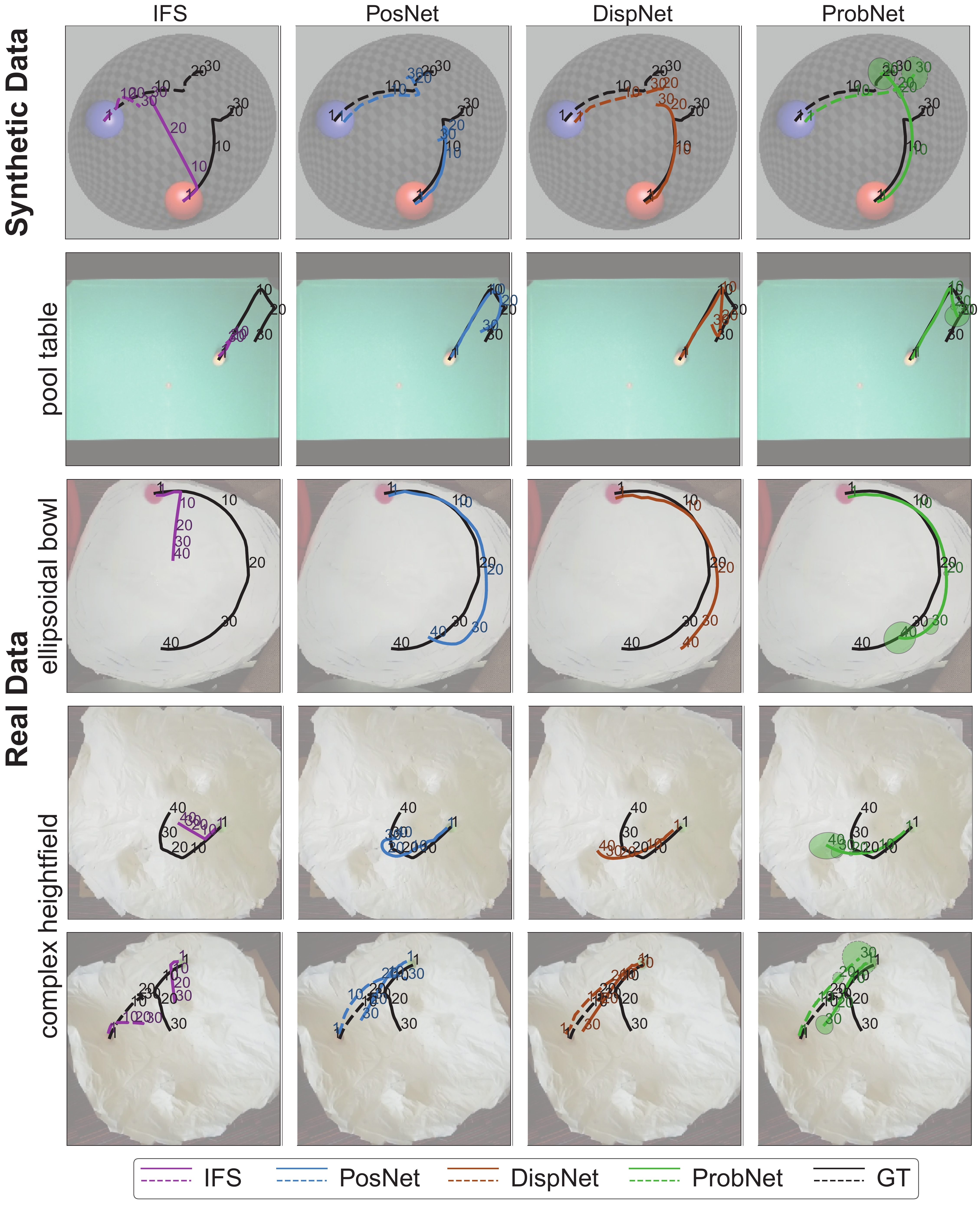} 
    \caption{\textbf{Qualitative performance comparison for the various methods against \groundtruth trajectories} Top-to-bottom: two balls colliding on an ellipsoidal bowl; single ball colliding against the walls of a pool table; single ball rolling on an ellipsoidal bow; single ball rolling on complex \heightfield; and two balls rolling on complex \heightfield. The top row is on synthetic data, while the other rows are on real-data. The green ellipsoids in the last column show the variance of the predictions estimated by \PNet at selected locations.}\label{f:results_plate}
\end{figure}

\subsection{Unsupervised Physics Extrapolation}

\paragraph{Supervised vs Unsupervised (Single Ball Synthetic Datasets).} We now compare training predictors using either \groundtruth object positions or the output of the unsupervised tracker. All predictors observe only $T_0 = 4$ frames as input (either positions or video frames) except VIN which uses $T_0 = 6$ and the least squares baselines which use $T_0 = 10$. All the networks were trained to predict $T_{train}$ positions. \Cref{t:one-real} reports the average errors at time $T_{train}$ and $2 T_{train}$ to measure the ability of predictors to generalise beyond the training regime.

We see that the \netmod (\PNet, \TNet, \PTNet) perform well using ground-truth positions or the unsupervised tracker outputs (e.g.\ \PTNet error for \bowls/\heightFs is 2.9/6.4 supervised vs 4.9/6.9 unsupervised), whereas IFS does not handle the transition well (3.3/10.4 to 13.3/23.1) and Linear, Quadratic and VIN are not competitive. The latest result shows a clear advantage of tensor-based state representations compared to vector based one. This suggests that modelling objects positions is done better by a representation which is spatially distributed. IFS also seems very sensitive to defects in the supplied annotations, since its knowledge of the environment is very limited, error correction is very challenging for it.

\input{tab-one-synth} 

The main weakness of the \netmod is that their performance degrades as prediction extends beyond the training horizon $2 T_{train}$, whereas IFS generalizes more. At least \PNet explicitly indicates that the model is uncertain when this occurs.

\paragraph{Synthetic vs Real (One Ball Datasets).} On real datasets (\cref{t:one-real}), the \netmod uniformly outperform others at both $T_{train}$ and $2 T_{train}$, with errors comparable to the synthetic case. Note that the real datasets in \benchmark are particularly challenging due to the non-idealities of the surface (e.g.\ the \bowl surface is slightly elastic and wobbles as the ball rolls).

\input{tab-one-real}

\paragraph{One vs Multiple Balls (Real and Synthetic Datasets).} Finally, we move to cases where the balls are interacting with the environment and with each others due to collisions. This is particularly challenging when no \groundtruth is used as multiple object tracking is much harder to achieve in an unsupervised setting than tracking a single object.


\input{tab-two} 

As shown in~\cref{t:two}, the \netmod still perform well. Due to memory limitations, models were trained for a slightly shorter time span $T_{train}$; since the corresponding predictions are shorter term, their errors are a little lower than before. Overall, the results show that neither perfect \groundtruth annotations nor a very large dataset is required to train a reliable physical extrapolator.
Still, we noticed that collisions were  difficult to predict in the \heightF dataset (see the bottom row of~\cref{f:results_plate}), probably because such events are rare during training. In contrast, this seems to be much better handled by the models in the synthetic dataset (First row of~\cref{f:results_plate}).

\subsection{Unsupervised Physics Interpolation}\label{sec:interp}

As in~\cite{LearningMechanics:emvm:2017}, we also study the interpolation problem considering their \IntNet configuration. We compare the latter to the extrapolation network \TNet trained over a longer horizon $T_{train}=\{30,40\}$. \IntNet has the same architecture has \TNet with the difference that, in addition to the first $T_0$ frames of the sequence, \IntNet additionally takes as input the last video frame as well. The first extracted state is used to regress the first $T_0$ positions as well as the positions at time $T_{train}$, so that this state is explicitly encouraged to encode information about the last position of the object as well.
In~\cref{t:interp} and \cref{t:interp2} we see that \IntNet managed to reduce the error in most cases. However in this case, compared to results in~\cite{LearningMechanics:emvm:2017} \IntNet performs poorer on synthetic dataset and estimation of the intermediate states seems to be more challenging.
Our interpretation is that the imperfect nature of the training data creates several possible path that this model in unable to solve.
Finally we also noticed that the heightfield datasets seem to be very challenging as training for longer horizons didn't reduce the error as much as it does on the `bowl.' 

\input{tab-interp}

%% file: fig-bartracker.tex
\begin{figure*}[h!]%
\includegraphics[width=0.61\columnwidth]{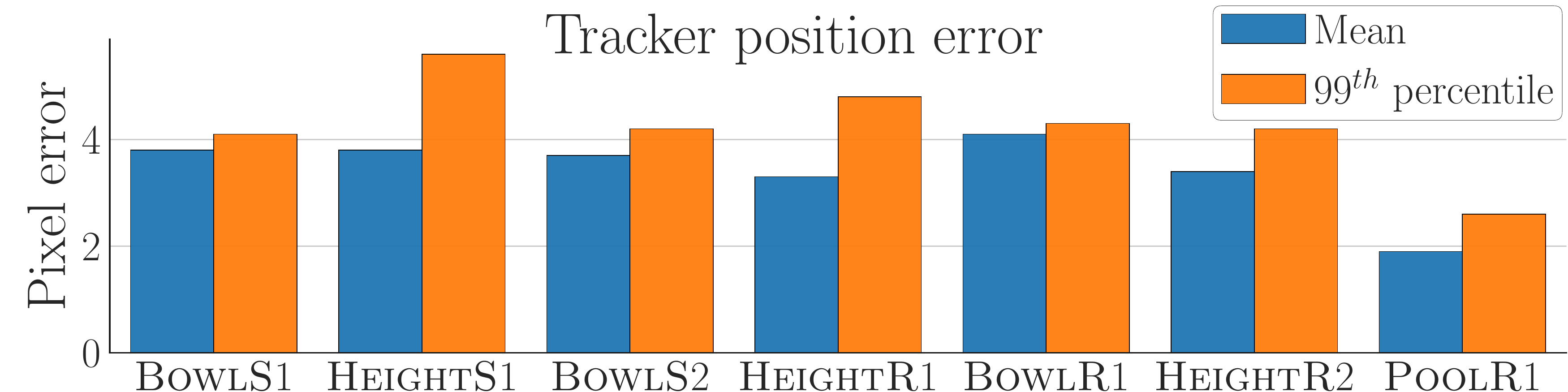}
\includegraphics[width=0.385\columnwidth]{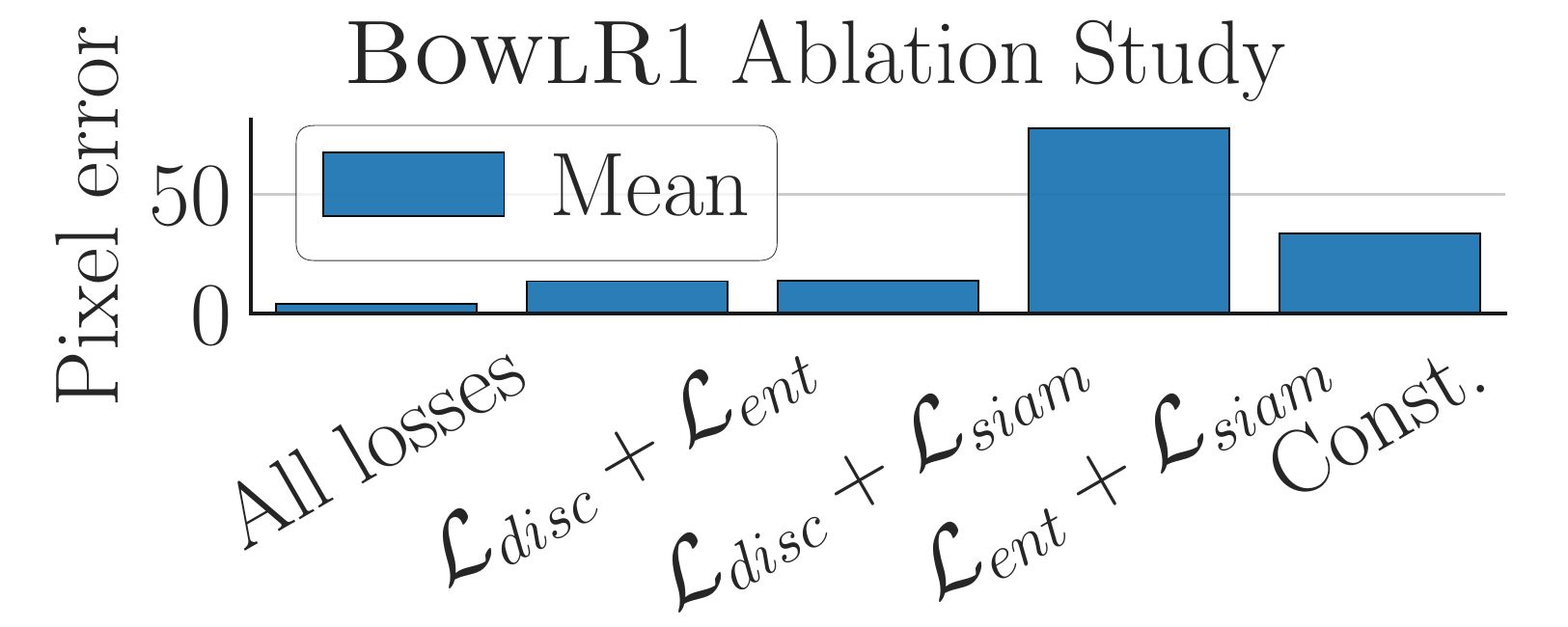}
    \caption{\textbf{Tracker errors and Ablation study} Left: Tracker errors on different dataset. The errors are consistently small across dataset and show that our tracker can perform well on a different range of real situations. Right: Ablation study. We try different combination of tracker losses on the \bowl dataset. `Const.' indicates that we are predicting a constant point at the center of the image for reference. For left and right, position errors are reported in pixels. The number of balls in the datasets is appended to the name of the dataset.}%
    \label{fig:bars-tracker}
\end{figure*}%

%% file: tab-comp-tracker.tex
\begin{table}[h]
    \centering
    \caption{\textbf{Tracker results across real datasets} The reported numbers are the average (left) and the variance (right) of the pixel error.  All numbers refer to 128$\times$128 images.}
    \footnotesize
    \begin{tabular}{lrrrrrrrr}
    \toprule
    &  \multicolumn{2}{c}{\pool} & \multicolumn{2}{c}{\bowl}  & \multicolumn{2}{c}{\heightF}   & \multicolumn{2}{c}{\heightF2B.} \\
        \midrule
        1. Optical Flow Lucas-Kanade & 23.3 & 965 &  5.6  & 275 &  \bfseries 2.7 & 12.9 & \bfseries 2.0  & 5.3  \\
        2a. FlowNet2-simple           & 41.4 &767 & 30.4 & 715 &  16.6 & 206 & - & -     \\
        2b. FlowNet2-blob             &  3.9  & 12.1 & \bfseries 2.2 & 4.8 &  4.6 & 28.7 & - & -   \\
        3. LAB w/o manual correction & \bfseries 0.3 & \bfseries 0.1 & 16.4 & 247 & 8.3 & 104 & 21.7 & 102\\
        4. Ours                      & 1.9 &  0.2 & 4.1 & \bfseries 0.5 & 3.3 & \bfseries 0.5 &3.4 & \bfseries 1.2\\
    \bottomrule
    \end{tabular}
    \label{t:results}
\end{table}

%% file: tab-one-synth.tex
\begin{table*}[t]
\centering
\caption{\textbf{Long term predictions compared on synthetic datasets with model trained with \groundtruth from simulator}  All the models (except VIN, Linear, and Quadratic) are given $T_0 = 4$ frames as input and train to predict first $T_{train}$ positions. We report the average pixel error and perplexity for \PTNet model at two different times. Perplexity, shown in bracket, is defined as $2^{-\mathbb{E}[\log_2(p(x))]}$ where $p$ is the estimated posterior distribution. \textit{State} shows either the carried forward state is a physical quantity (\expl), or an implicit vector or tensor (\impl)}

\scriptsize
\sisetup{detect-weight=true,detect-inline-weight=math,  table-column-width=6.5em}
\newcommand{\boldentry}[2]{%
\multicolumn{1}{S[table-format=0.1,
                    mode=text, text-rm=\fontseries{b}\selectfont
                   ]#2}{#1}}
\newcolumntype{L}{S[table-format=0.1]}
\begin{tabular}{ccc*2L*2L}
\toprule
    & && \multicolumn{2}{c}{\bowltwo - \Ttrain = 20}  & \multicolumn{2}{c}{\heightFs - \Ttrain = 20} \\
    \cmidrule(lr){4-5} \cmidrule(lr){6-7}
    & &
    & \multicolumn{1}{c}{$T=$\Ttrain}
    & \multicolumn{1}{c}{$2\times$\Ttrain}
    & \multicolumn{1}{c}{\Ttrain}
    & \multicolumn{1}{c}{$2\times$\Ttrain}\\[0.1cm]
    \midrule
    Method &  Input & State &
    \multicolumn{4}{c}{With positions from simulator}\\
    \midrule
    Linear & 2D pos. & \expl &  61.9 &  20.1 &  21.3 &  61.9\\
    Quadratic & 2D pos. & \expl &  11.7 &  93.1 &   26.7 &  126.0 \\
    IFS & 2D pos. & \expl &  3.3  & \boldentry{8.9}{} & 10.4  & 27.6  \\
    VIN & Visual & \impl & 24.0  & 30.2  & 42.6   & 42.7 \\
    \midrule
    \PTNet & Visual & \impl  &  \boldentry{1.6}{}  & 24.4  & 7.2   & 24.6 \\
    \TNet & Visual & \impl &  2.5  & 20.6  &  7.7   & 25.8 \\
    \PNet & Visual & \impl  & 2.9 { (32.1)} & 21.8 { (54.0)}   &  \boldentry{6.4 \textsc{{ (9.5 )}}}{} &  \boldentry{22.5 \text{{ (12.7)} }}{} \\
    \midrule
    Method &  Input & State &
    \multicolumn{4}{c}{With positions from \textit{unsupervised} tracker} \\
    \midrule
    IFS & 2D pos. & \expl &  13.3 & \boldentry{23.6}{}  &  23.1 & 38.3  \\
    VIN & Visual & \impl & 24.7  & 30.3  & 45.8   & 48.0 \\
    \midrule
    \PTNet & Visual & \impl  &  4.3 & 29.9  & \boldentry{6.6}{}  & 25.6 \\
    \TNet & Visual & \impl &  \boldentry{3.9}{} & 25.6  &  6.8 & \boldentry{22.7}{} \\
    \PNet & Visual & \impl  & 4.9 { (6.3)} & 27.0 { (20.6)}  &  6.9 { (8.3 )} &  23.3 { (13.4)}\\
\bottomrule
\end{tabular}
\label{t:one}
\end{table*}

%% file: tab-one-real.tex
\begin{table*}[t]
\centering
\caption{\textbf{Long term predictions using one ball and real data}
The table has the same format as~\cref{t:one}. All models are trained using the unsupervised tracker, input and state are the same as~\cref{t:one}, and we report pixel error (perplexity) at $T$.}\label{t:one-real}

\scriptsize
\sisetup{detect-weight=true,detect-inline-weight=math,  table-column-width=5.0em}
\newcommand{\boldentryy}[2]{%
\multicolumn{1}{S[table-format=0.1,
                    mode=text, text-rm=\fontseries{b}\selectfont
                   ]#2}{#1}}
\newcolumntype{F}{S[table-format=0.1]}
\begin{tabular}{c*2F*2F*2F}
\toprule
    & \multicolumn{2}{c}{\pool-$T_{train} = 15$}
    & \multicolumn{2}{c}{\heightF - $T_{train} = 20$}
    & \multicolumn{2}{c}{\bowl- $T_{train} = 20$} \\
    \cmidrule(lr){2-3} \cmidrule(lr){4-5} \cmidrule(lr){6-7}
    Method &
    \multicolumn{1}{c}{$T=$\Ttrain} &
    \multicolumn{1}{c}{$2\times$\Ttrain} & \multicolumn{1}{c}{\Ttrain} &
    \multicolumn{1}{c}{$2\times$\Ttrain} & \multicolumn{1}{c}{\Ttrain} &
    \multicolumn{1}{c}{$2\times$\Ttrain}\\[0.1cm]

    \midrule
    IFS &  26.0 & 37.5 &   48.0  & 58.1  & 26.2  & 39.1  \\
    VIN &  50.9 & 40.8 & 40.2  & 47.3  & 33.9   & 33.0 \\
    \midrule
    \PTNet & 4.6 & 21.4 & \boldentryy{5.6}{} & 29.0  &  \boldentryy{5.6}{} & 23.0 \\
    \TNet &  \boldentryy{3.8}{} & 23.6 & \boldentryy{5.6}{}  & \boldentryy{28.5}{}  &  6.5  & \boldentryy{22.6}{} \\
    \PNet & 4.7 {(6.3)}&  \boldentryy{16.3 \textsc{{(11.3)}}}{}  &  5.7{(5.8)} & 30.0{(22.5)}   &  6.8{(6.8)} &  23.5{(13.8)} \\
\bottomrule
\end{tabular}
\end{table*}

%% file: tab-two.tex
\begin{table*}[t]
\centering
\caption{\textbf{Long term predictions using two balls on real and synthetic data} Table layout and measures are the same as \cref{t:one}. Models are trained with positions from tracker, input and state are the same as~\cref{t:one}, and we report pixel error (perplexity) at $T$.}\label{t:two}
\sisetup{detect-weight=true,detect-inline-weight=math,  table-column-width=6.7em}
\newcommand{\boldentryp}[2]{%
\multicolumn{1}{S[table-format=0.4,
                    mode=text, text-rm=\fontseries{b}\selectfont
                   ]#2}{#1}}
\newcolumntype{T}{S[table-format=0.4]}
\begin{tabular}{c*2T*2T}
\toprule
    Method
    & \multicolumn{2}{c}{\bowltwo  2b.-$T_{train}=15$} & \multicolumn{2}{c}{\heightF 2b.-$T_{train}=15$}\\
    \cmidrule(lr){2-3} \cmidrule(lr){4-5}
    &
    \multicolumn{1}{c}{$T=$\Ttrain} &
    \multicolumn{1}{c}{$2\times$\Ttrain} &
    \multicolumn{1}{c}{\Ttrain} &
    \multicolumn{1}{c}{$2\times$\Ttrain}\\
    \midrule
    IFS &   18.4  & 30.0  & 15.6  & 26.6  \\
    VIN &   41.3 & 45.8  & 45.9  & 39.8 \\
    \midrule
    \PTNet &  \boldentryp{5.0}{} & \boldentryp{13.4}{} &  \boldentryp{5.4}{} & \boldentryp{12.5}{} \\
    \TNet &           5.5 & 24.7            &  6.2         & 15.4 \\
    \PNet &   5.6 {(7.3)} & 20.6  {(13.7)}  &  6.8 {(7.9)} &  16.9 {(12.4)}\\
    \bottomrule
\end{tabular}
\end{table*}

%% file: tab-interp.tex
\begin{table*}[t]
\setlength{\tabcolsep}{0.35em}
\newcolumntype{K}{S[table-format=0.1, table-column-width=1.2em]}
\centering
\caption{\textbf{Extrapolation vs interpolation: one ball datasets} One ball datasets synthetic and real. Models are trained with positions from tracker. Pixel error at different time $T$.}
\scriptsize
\sisetup{detect-weight=true,detect-inline-weight=math, table-column-width=1.2em}
\begin{tabular}{c*3K*4K*4K*4K*4K}
\toprule
& \multicolumn{3}{c}{\pool}
& \multicolumn{4}{c}{\bowltwo} 
& \multicolumn{4}{c}{\heightFs}
& \multicolumn{4}{c}{\bowl}
& \multicolumn{4}{c}{\heightF}\\ 
& \multicolumn{3}{c}{$T_\text{train}=30$}
& \multicolumn{4}{c}{$T_\text{train}=40$} 
& \multicolumn{4}{c}{$T_\text{train}=40$} 
& \multicolumn{4}{c}{$T_\text{train}=40$}
& \multicolumn{4}{c}{$T_\text{train}=40$} \\
\cmidrule(lr){2-4} \cmidrule(lr){5-8}
\cmidrule(lr){9-12} \cmidrule(lr){13-16}
\cmidrule(lr){17-20}
$T$
& {10} & {20} & {30}
& {10} & {20} & {30} & {40}
& {10} & {20} & {30} & {40}
& {10} & {20} & {30} & {40}
& {10} & {20} & {30} & {40}\\
\midrule 
\TNet & 3.1 & 5.6 & 10.1 & 3.8 & 4.0 & 4.2 & 4.2 & 5.2 & 8.2 & 13.2 & 19.1 & 4.1 & 5.0 & 5.5 & 6.9 & 4.3 & 6.6 & 9.4 & 12.7 \\
\IntNet & 4.5 & 5.6 & 3.1 & 3.8 & 4.2 & 4.0 & 3.8 & 4.5 & 6.4 & 6.5 & 4.2 & 6.3 & 6.5 & 4.8 & 3.7  & 4.0 & 5.0 & 4.8 & 4.3 \\
\bottomrule    
\end{tabular}
\label{t:interp}
\end{table*}

\begin{table*}[t]
\setlength{\tabcolsep}{1.2em}
\newcolumntype{K}{S[table-format=0.1, table-column-width=2.5em]}
\centering
\caption{\textbf{Extrapolation vs interpolation: two balls datasets} Two balls datasets synthetic and real. Models are trained with positions from tracker. Pixel error at different time $T$.}
\sisetup{detect-weight=true,detect-inline-weight=math, table-column-width=1.4em}
\begin{tabular}{c*3K*3K}
\toprule
& \multicolumn{3}{c}{\bowltwo2b} & \multicolumn{3}{c}{\heightF2b} \\ 
& \multicolumn{3}{c}{$T_\text{train}=30$} & \multicolumn{3}{c}{$T_\text{train}=30$}  \\
\cmidrule(lr){2-4} \cmidrule(lr){5-7}
$T$
& {10} & {20} & {30}
& {10} & {20} & {30}
\\
\midrule 
\TNet & 4.3 & 6.9 & 9.7 & 5.2 & 8.9 & 13.4 \\
\IntNet & 4.2 & 5.0 & 4.1 & 6.5 & 6.9 & 7.6 \\
\bottomrule    
\end{tabular}\label{t:interp2}
\end{table*}

%% file: conclusions.tex
\section{Conclusions}\label{s:conclusions}

We presented a method that can learn to track physical objects such as  balls rolling on complex terrains using only raw video sequences and no supervision. Combined with recent neural networks that can learn an implicit representation of physics, such a system is able to  extrapolate object trajectories over time while accounting for object-environment and object-object interactions. To the best of our knowledge, this is the first time that learning long-term physics extrapolation without access to supervision or handcrafted simulators has been demonstrated. Through an extensive benchmark we also demonstrated the superiority of tensor-based state representation that were able to produce satisfactory results on real data without the need of large datasets.

We also contributed a new dataset, \benchmark, of real-life video sequences for complex scenarios such as ball rollings on pool tables, bowls, and \heightfield, showing that all such methods are applicable to the real world. This data will be made publicly available. 

In this work we used different colored objects to make them distinguishable, which in practice is one of the main limitation of our work. We plan to address this issue by using same colored objects and build a tracker that would be trained to detect all objects at once removing the need for iterative training. 

Finally, we also plan to train the tracker and the extrapolator end-to-end, further improving tracking of multiple objects. 
We also aim at improving the generalisation of the predictors beyond the training regime; we believe that the key is to factor knowledge about the environment and the object dynamics to allow the models to remember the first better over longer time spans.

{\bf Acknowledgements.} The authors would like to gratefully acknowledge the support of ERC 677195-IDIU and ERC SmartGeometry StG-2013-335373 grants.

%% file: main.bbl
\begin{thebibliography}{10}
\providecommand{\url}[1]{\texttt{#1}}
\providecommand{\urlprefix}{URL }
\providecommand{\doi}[1]{https://doi.org/#1}

\bibitem{tensorflow2015:whitepaper}
Abadi, et~al.: {TensorFlow}: Large-scale machine learning on heterogeneous
  systems (2015), software available from tensorflow.org

\bibitem{NIPS2016_6113}
Agrawal, P., et~al.: {Learning to Poke by Poking: Experiential Learning of
  Intuitive Physics}. In: Proc. {NIPS}. pp. 5074--5082 (2016)

\bibitem{battaglia2016interaction}
Battaglia, P., et~al.: Interaction networks for learning about objects,
  relations and physics. In: Proc. {NIPS}. pp. 4502--4510 (2016)

\bibitem{battaglia2013simulation}
Battaglia, P., Hamrick, J., Tenenbaum, J.: Simulation as an engine of physical
  scene understanding. {PNAS}  \textbf{110}(45),  18327--18332 (2013)

\bibitem{bhattacharyya2018long}
Bhattacharyya, A., et~al.: Long-term image boundary prediction. In:
  Thirty-Second AAAI Conference on Artificial Intelligence. AAAI (2018)

\bibitem{opencv_library}
Bradski, G.: {The OpenCV Library}. Dr. Dobb's Journal of Software Tools  (2000)

\bibitem{chang2016compositional}
Chang, M.B., et~al.: A compositional object-based approach to learning physical
  dynamics. In: Proc. {ICLR} (2017)

\bibitem{recurrentenv}
Chiappa, S., et~al.: Recurrent environment simulators (2017)

\bibitem{denil2016learning}
Denil, M., et~al.: Learning to perform physics experiments via deep
  reinforcement learning. Deep Reinforcement Learning Workshop, {NIPS}  (2016)

\bibitem{LearningPhysicalPredictor:emmv:2017}
Ehrhardt, S., others.: {Learning A Physical Long-term Predictor}. arXiv
  e-prints arXiv:1703.00247  (Mar 2017)

\bibitem{LearningMechanics:emvm:2017}
Ehrhardt, S., et~al.: {Learning to Represent Mechanics via Long-term
  Extrapolation and Interpolation}. arXiv preprint arXiv:1706.02179  (Jun 2017)

\bibitem{eslami2016attend}
Eslami, S.A., et~al.: Attend, infer, repeat: Fast scene understanding with
  generative models. In: Advances in Neural Information Processing Systems. pp.
  3225--3233 (2016)

\bibitem{finn2016deep}
Finn, C., et~al.: Deep spatial autoencoders for visuomotor learning. In:
  Robotics and Automation (ICRA), 2016 IEEE International Conference on. pp.
  512--519. IEEE (2016)

\bibitem{fragkiadaki16learning}
Fragkiadaki, K., et~al.: Learning visual predictive models of physics for
  playing billiards. In: Proc. {NIPS} (2016)

\bibitem{glorot2010understanding}
Glorot, X., Bengio, Y.: Understanding the difficulty of training deep
  feedforward neural networks. In: Proceedings of the thirteenth international
  conference on artificial intelligence and statistics. pp. 249--256 (2010)

\bibitem{ilg2017flownet}
Ilg, E., Mayer, N., Saikia, T., Keuper, M., Dosovitskiy, A., Brox, T.: Flownet
  2.0: Evolution of optical flow estimation with deep networks

\bibitem{kansky2017schema}
Kansky, K., et~al.: Schema networks: Zero-shot transfer with a generative
  causal model of intuitive physics. In: International Conference on Machine
  Learning. pp. 1809--1818 (2017)

\bibitem{jeong2015data}
Ladický, et~al.: Data-driven fluid simulations using regression forests. ACM
  Trans. on Graphics (TOG)  \textbf{34}(6), ~199 (2015)

\bibitem{lee2018savp}
Lee, A.X., et~al.: Stochastic adversarial video prediction. arXiv preprint
  arXiv:1804.01523  (2018)

\bibitem{lerer2016learning}
Lerer, A., Gross, S., Fergus, R.: Learning physical intuition of block towers
  by example. In: Proceedings of the 33rd International Conference on
  International Conference on Machine Learning - Volume 48. pp. 430--438 (2016)

\bibitem{li2016visual}
Li, W., Leonardis, A., Fritz, M.: Visual stability prediction and its
  application to manipulation. AAAI  (2017)

\bibitem{neverova2017predicting}
Luc, P., Neverova, N., Couprie, C., Verbeek, J., LeCun, Y.: Predicting deeper
  into the future of semantic segmentation. ICCV  (2017)

\bibitem{misra2016shuffle}
Misra, I., Zitnick, C.L., Hebert, M.: Shuffle and learn: unsupervised learning
  using temporal order verification. In: European Conference on Computer
  Vision. pp. 527--544. Springer (2016)

\bibitem{MonszpartEtAl:SMASH:2016}
Monszpart, A., Thuerey, N., Mitra, N.: {SMASH: Physics-guided Reconstruction of
  Collisions from Videos}. ACM Trans. on Graphics (TOG)  (2016)

\bibitem{Mottaghi_2016_CVPR}
Mottaghi, R., et~al.: Newtonian scene understanding: Unfolding the dynamics of
  objects in static images. In: {IEEE CVPR} (2016)

\bibitem{2018arXiv06}
Mrowca, D., et~al.: {Flexible Neural Representation for Physics Prediction}.
  ArXiv e-prints  (2018)

\bibitem{novotny2018self}
Novotny, D., et~al.: Self-supervised learning of geometrically stable features
  through probabilistic introspection (2018)

\bibitem{oh2015action}
Oh, J., et~al.: Action-conditional video prediction using deep networks in
  atari games. In: Advances in Neural Information Processing Systems. pp.
  2863--2871 (2015)

\bibitem{OndruskaAAAI2016}
Ondruska, P., Posner, I.: Deep tracking: Seeing beyond seeing using recurrent
  neural networks. In: Proc. {AAAI} (2016)

\bibitem{2018arXiv180307616R}
{Riochet}, R., et~al.: {IntPhys: A Framework and Benchmark for Visual Intuitive
  Physics Reasoning}. ArXiv e-prints  (2018)

\bibitem{sanborn2013reconciling}
Sanborn, A.N., Mansinghka, V.K., Griffiths, T.L.: Reconciling intuitive physics
  and newtonian mechanics for colliding objects. Psychological review
  \textbf{120}(2), ~411 (2013)

\bibitem{sanchez2018graph}
Sanchez-Gonzalez, A., et~al.: Graph networks as learnable physics engines for
  inference and control  (2018)

\bibitem{Stewart2016LabelFreeSO}
Stewart, R., Ermon, S.: Label-free supervision of neural networks with physics
  and domain knowledge. In: AAAI. pp. 2576--2582 (2017)

\bibitem{thewlis2017unsupervised}
Thewlis, J., Bilen, H., Vedaldi, A.: Unsupervised learning of object frames by
  dense equivariant image labelling. In: Advances in Neural Information
  Processing Systems (NIPS). pp. 844--855 (2017)

\bibitem{CNNFluid2016}
{Tompson}, J., et~al.: {Accelerating Eulerian Fluid Simulation With
  Convolutional Networks}. ArXiv e-print arXiv:1607.03597  (2016)

\bibitem{NIPS2017_7040}
Watters, N., et~al.: Visual interaction networks: Learning a physics simulator
  from video. In: Guyon, I., Luxburg, U.V., Bengio, S., Wallach, H., Fergus,
  R., Vishwanathan, S., Garnett, R. (eds.) Advances in Neural Information
  Processing Systems 30, pp. 4542--4550. Curran Associates, Inc. (2017)

\bibitem{Galileo:NIPS:2015}
Wu, J., et~al.: Galileo: Perceiving physical object properties by integrating a
  physics engine with deep learning. In: Proc. {NIPS}. pp. 127--135 (2015)

\bibitem{phys101}
Wu, J., et~al.: Physics 101: Learning physical object properties from unlabeled
  videos. In: Proc. {BMVC} (2016)

\bibitem{NIPS2017_6620}
Wu, J., et~al.: Learning to see physics via visual de-animation. In: Guyon, I.,
  et~al. (eds.) Advances in Neural Information Processing Systems (NIPS) 30,
  pp. 153--164. Curran Associates, Inc. (2017)

\bibitem{wu17learning}
Wu, J., et~al.: Learning to see physics via visual de-animation. In: Proc.
  {NIPS} (2017)

\end{thebibliography}
